\documentclass[letterpaper, 10 pt, conference]{ieeeconf}
\IEEEoverridecommandlockouts                              
\usepackage{comment}

\usepackage{algorithm}
\usepackage{algorithmic}
\usepackage{url}
\usepackage{amsmath}
\usepackage{cite}
\usepackage{epsfig}
\usepackage{epic,eepic,color}
\usepackage{times}
\usepackage{mathptmx}
\usepackage{multirow}
\usepackage{stkernel}

\usepackage{cases}

\usepackage{times}
\usepackage{multirow}

\definecolor{red}{rgb}{1,0,0}
\definecolor{green}{rgb}{0,1,0}
\definecolor{blue}{rgb}{0,0,1}
\definecolor{violet}{rgb}{1,0,1}
\definecolor{cyan}{cmyk}{1,0,0,0}
\definecolor{magenta}{cmyk}{0,1,0,0}
\definecolor{yellow}{cmyk}{0,0,1,0}

\definecolor{white}{rgb}{1,1,1}

\usepackage{jumoline}

\newcommand{\CO}[1]{}

\newcommand{\CommentOut}[1]{}

\newcommand{\noeditage}[1]{#1} \newcommand{\editage}[1]{}

\newcommand{\FIG}[3]{
\begin{minipage}[b]{#1cm}
\begin{center}
\includegraphics[width=#1cm]{#2}\\
{\scriptsize #3}
\end{center}
\end{minipage}
}

\newcommand{\FIGR}[3]{
\begin{minipage}[b]{#1cm}
\begin{center}
\includegraphics[angle=-90,clip,width=#1cm]{#2}
\\
{\scriptsize #3}
\vspace*{1mm}
\end{center}
\end{minipage}
}

\newcommand{\FIGpng}[5]{
\begin{minipage}[b]{#1cm}
\begin{center}
\includegraphics[bb=0 0 #4 #5, clip, width=#1cm]{#2}\vspace*{-1mm}\\
{\scriptsize #3}
\vspace*{1mm}
\end{center}
\end{minipage}
}

\newcommand{\figA}{
\begin{figure}[t]
  \begin{center}
\FIG{8}{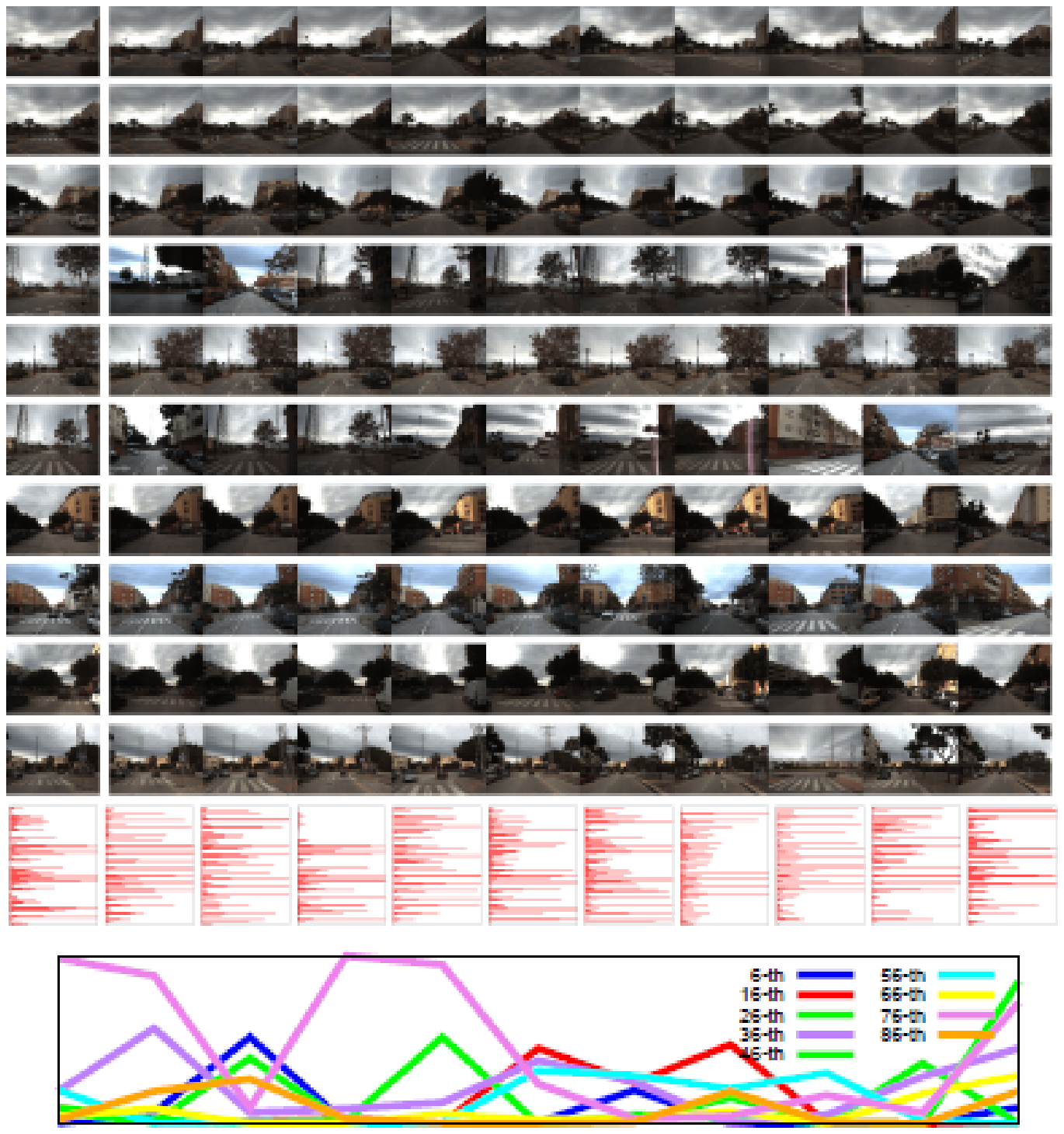}{}
\caption{Change detection under global viewpoint uncertainty.
Top:
Each row shows a query image (first column) and the top-ranked to 10-th ranked
reference images (from the second to the last column), respectively, in the map relative localization.
Middle:
Bag-of-local-convolutional-features (BoLCF) histogram for query and top-ranked reference images for the first row.
Bottom:
Visualization of a SIFT feature in the first query image and its nearest neighbor features in the top-ranked reference images.
Shown in the graph are values of several dimensions (6-th, 16-th, ..., 86-th dims) of the SIFT vectors.}\label{fig:A}
\end{center}
\end{figure}
}

\newcommand{\figB}{
\begin{figure}[t]
  \begin{center}
\FIG{7}{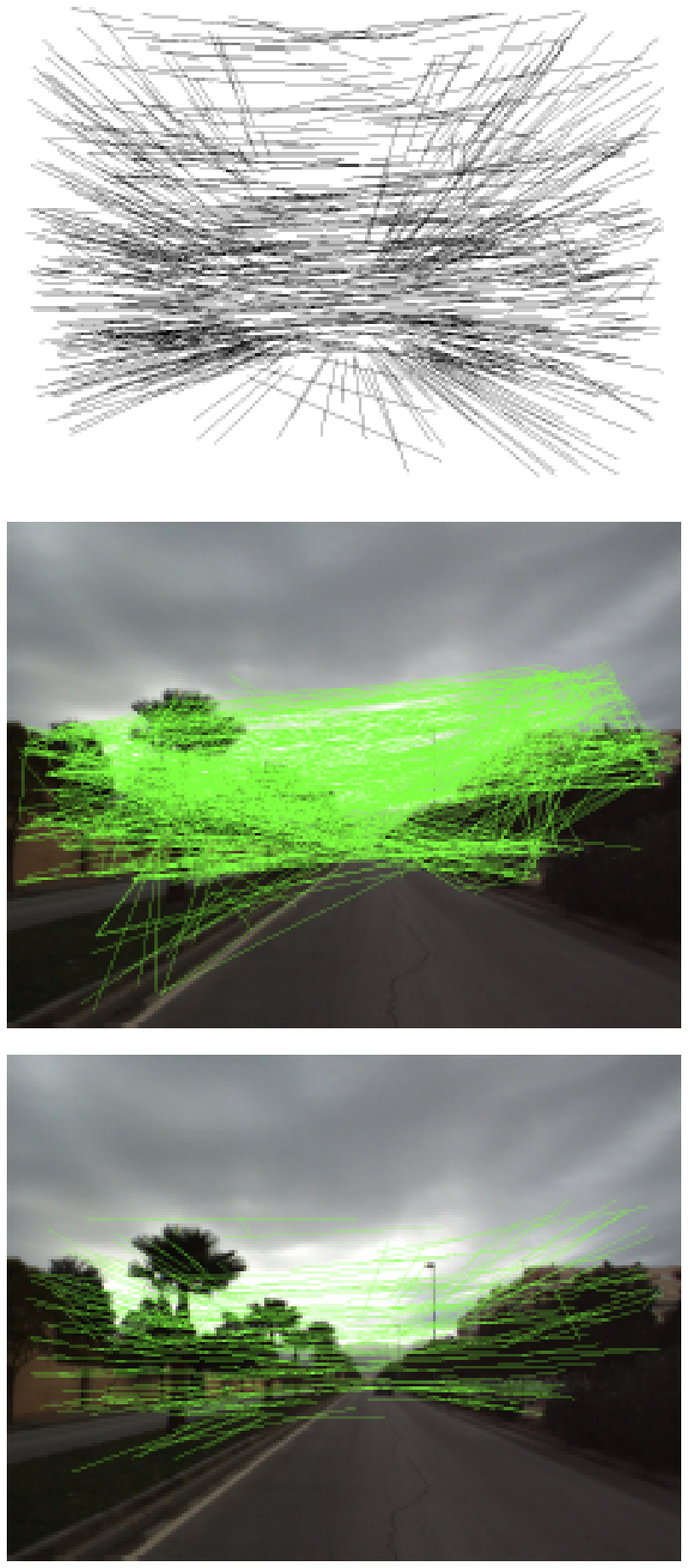}{}
\caption{Anomaly motion detection using motion prior. Top: Samples from the motion vocabulary learned from Malaga sequence \#9. Each line segment
corresponds to a motion word or a motion exemplar that is a 4D vector consisting of a pairing of 2D vectors, a vector at the start position and a vector at
the end position, of a feature track on the image plane. Middle: Each line segment indicates the motion between each keypoint in the query image (2D)
and its nearest neighbor keypoint in the reference image. Bottom: Nearest neighbor motion exemplars (4D vectors) explaining individual motion features.}\label{fig:B}
\end{center}
\vspace*{-7mm}
\end{figure}
}

\newcommand{\figC}{
\begin{figure}[t]
  \begin{center}
\FIGR{8}{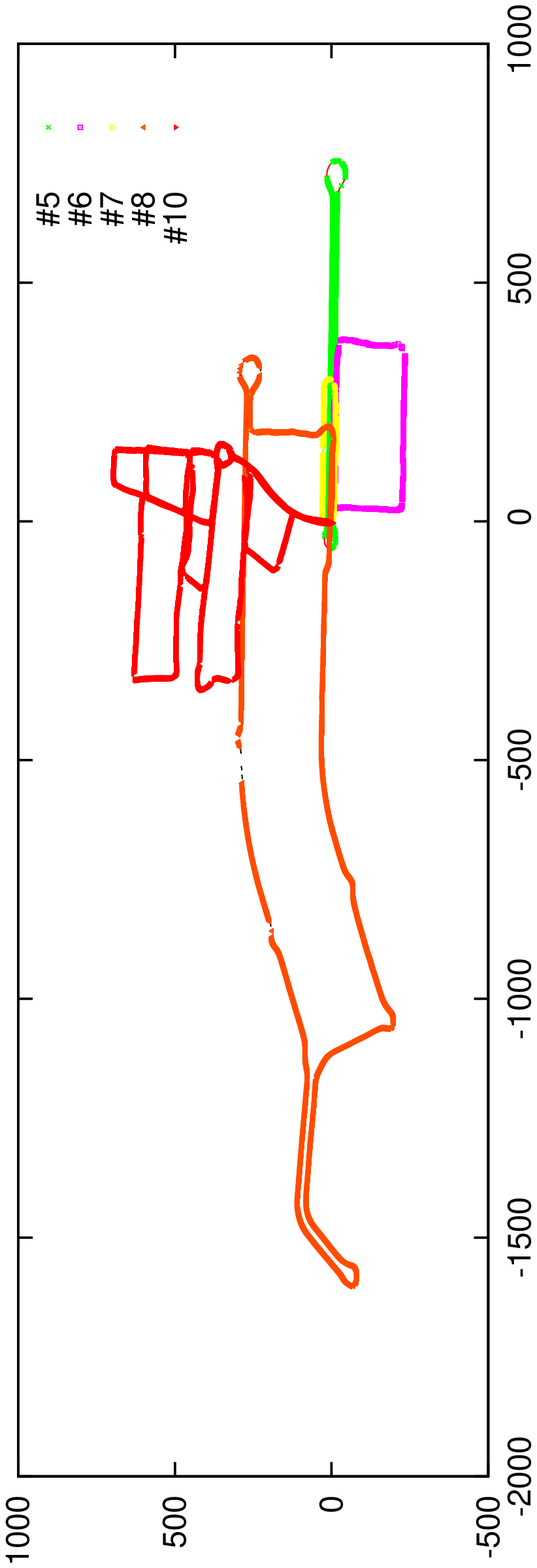}{}    
\caption{Bird's eye view of the experimental environments.
  Colored lines and points, respectively, indicate the robot's trajectories in [m] and the viewpoints where
the robot's ego-motions are recognized as non-anomaly ego-motions.}\label{fig:C}
\end{center}
\end{figure}
}

\newcommand{\figD}{
\begin{figure}[t]
  \begin{center}
\vspace*{-3mm}
\FIGR{9}{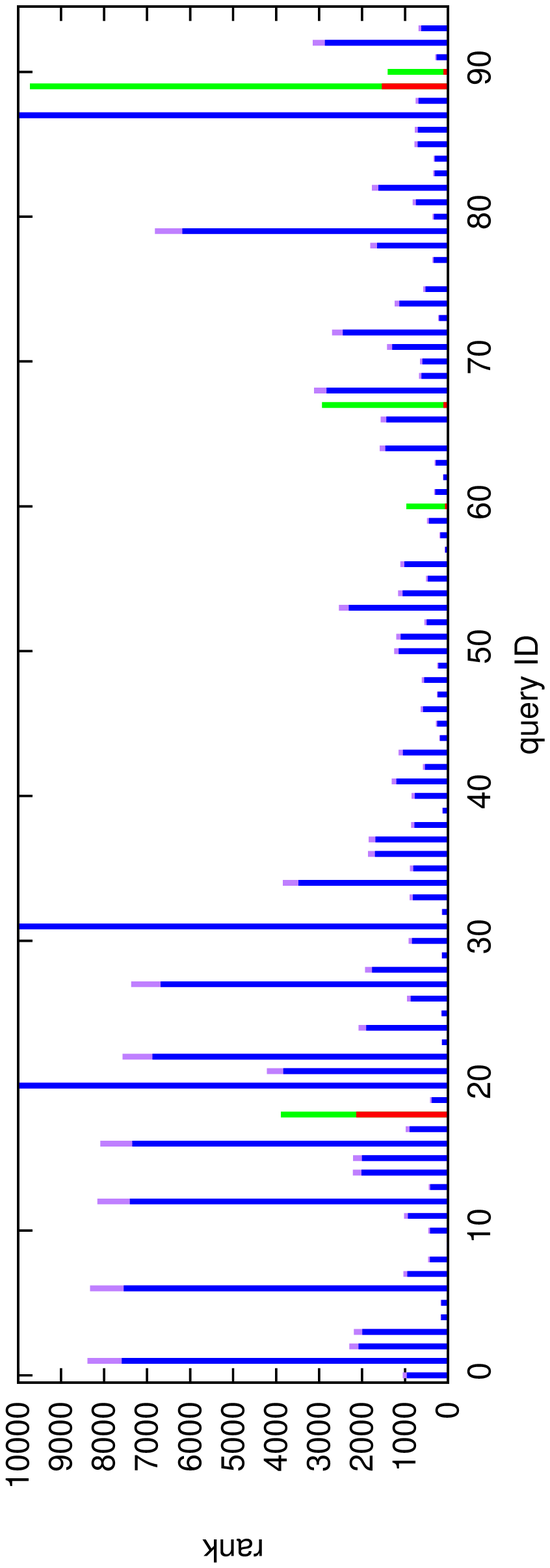}{}\vspace*{-3mm}\\
\hspace*{0.5mm}\FIGR{8.9}{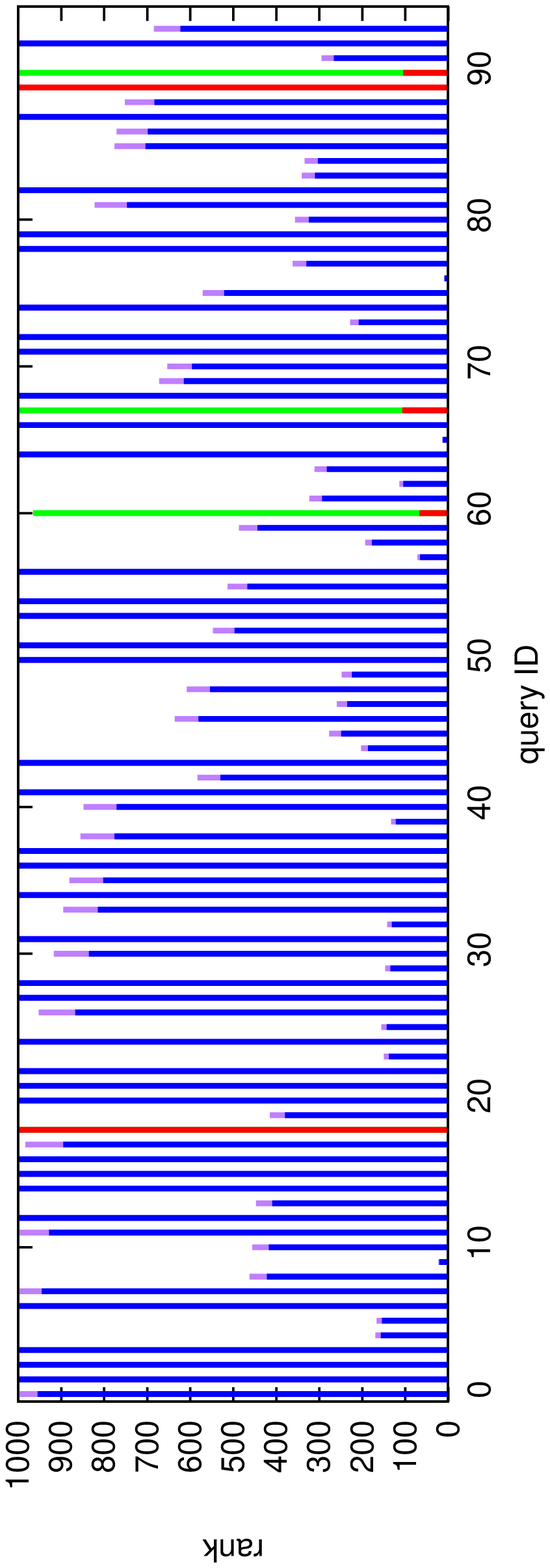}{}\vspace*{-3mm}\\
\caption{Individual change detection results. The length of the blue (or green) and purple (or red) line segments indicate the rank values or the likelihood of change estimated with and without using the motion prior. The blue and red colors respectively indicate the cases in which the former and the latter algorithms perform better than the other algorithm.}\label{fig:D}
\end{center}
\vspace*{-5mm}
\end{figure}
}

\newcommand{\figF}{
\begin{figure}[t]
\begin{center}
\FIGR{8.5}{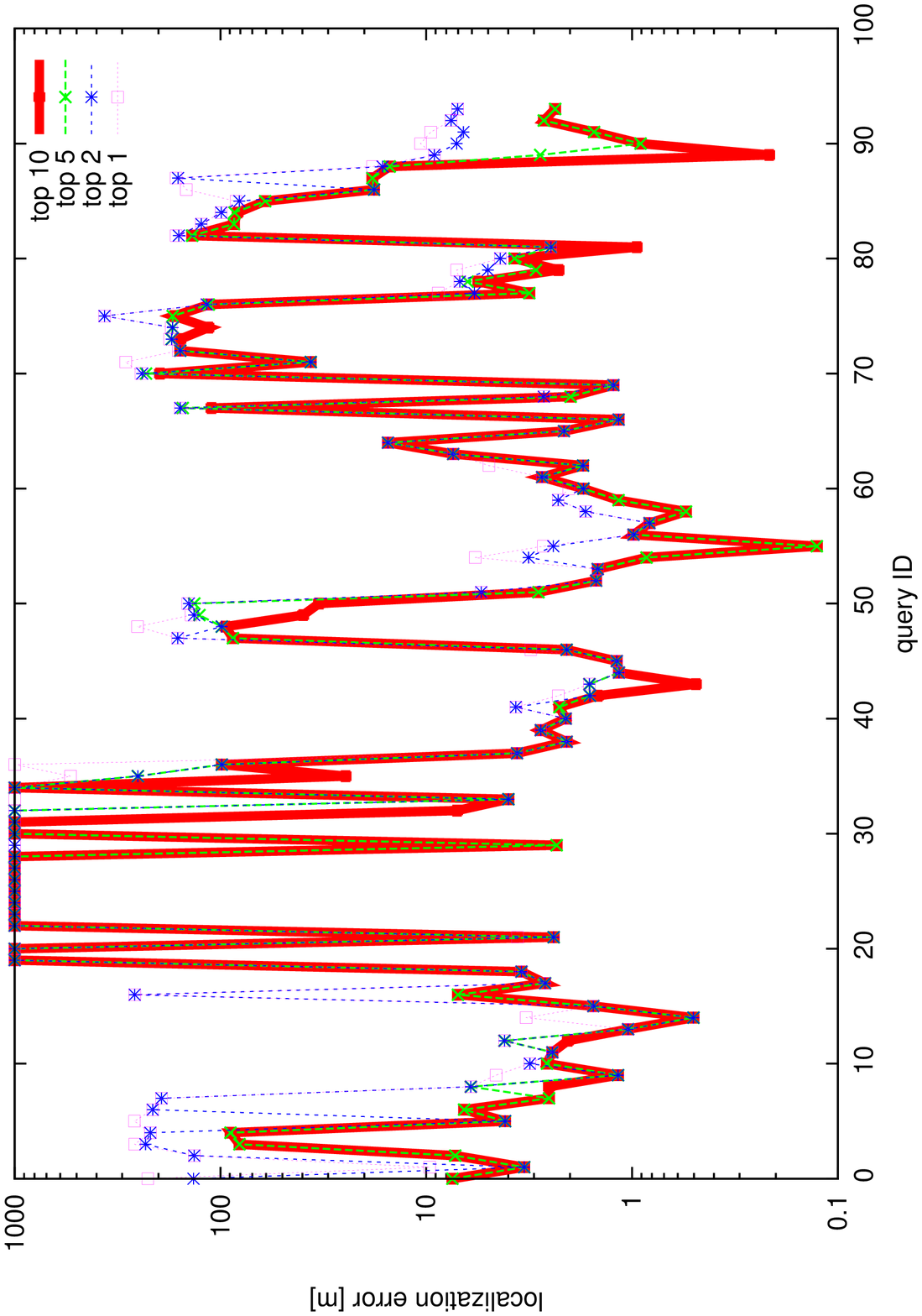}{(a)}\\
\FIGR{4}{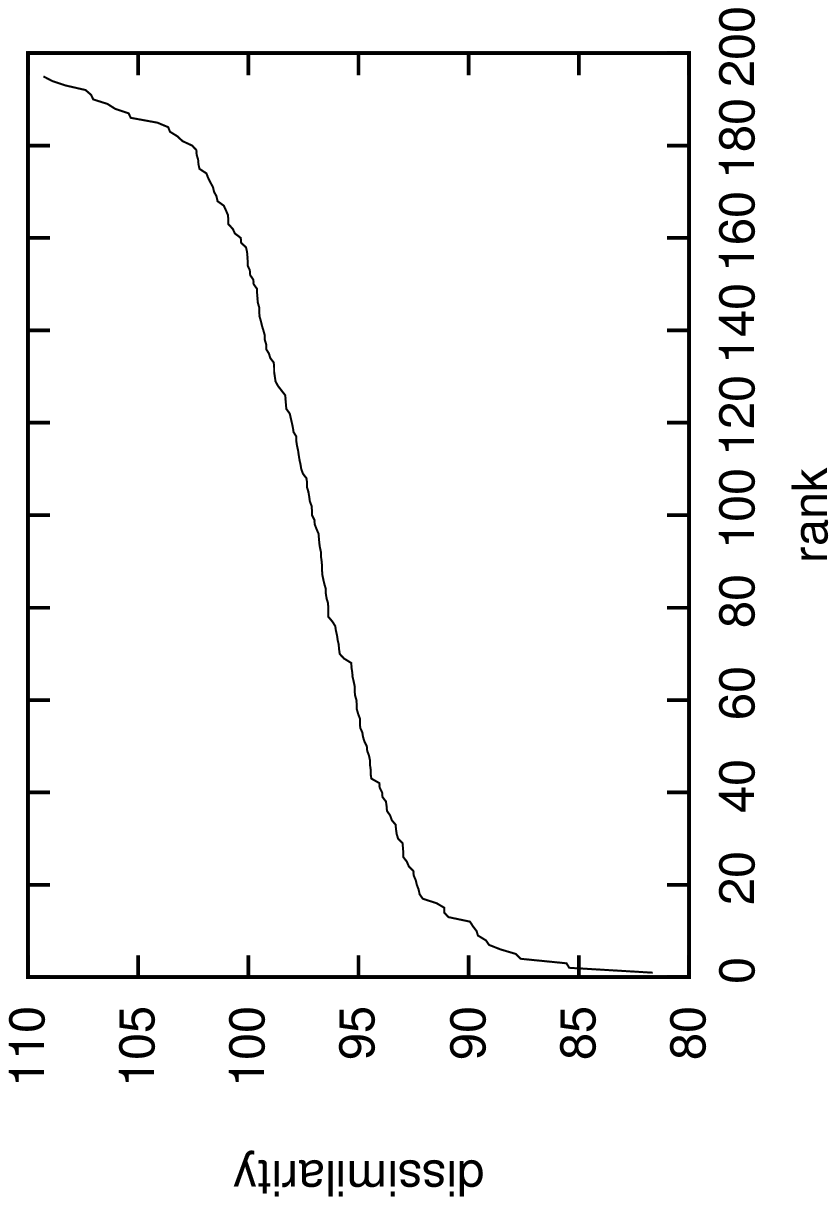}{}
\FIGR{4}{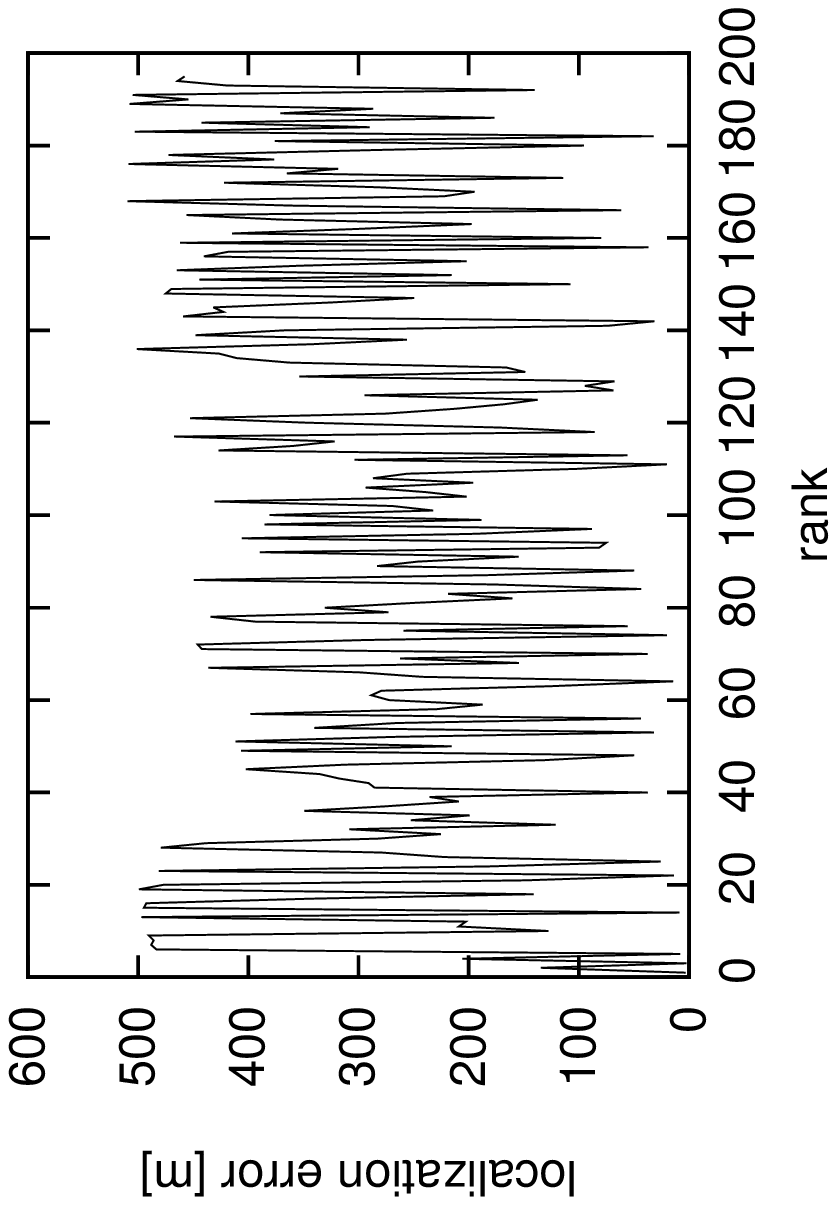}{}\\
{\scriptsize (b)}
\caption{Individual localization results.
  (a)
  Top-10, top-5, top-2, and top-1 recognition rates are shown in different colors.
  For the purpose of visualization, localization errors greater than 1,000 m are treated as 1,000 m in the graph.
  (b)
  Localization result for an example query image.
  Left:
  Dissimilarity of the BoLCF histogram between the query image and each $i$-th top-ranked reference image.
  Right:
Localization error in [m] for each $i$-th top-ranked reference image.
}\label{fig:F}
\end{center}
\end{figure}
}

\newcommand{\figG}{
\begin{figure}[t]
  \begin{center}
    \begin{minipage}[b]{8.7cm}
\FIGR{4}{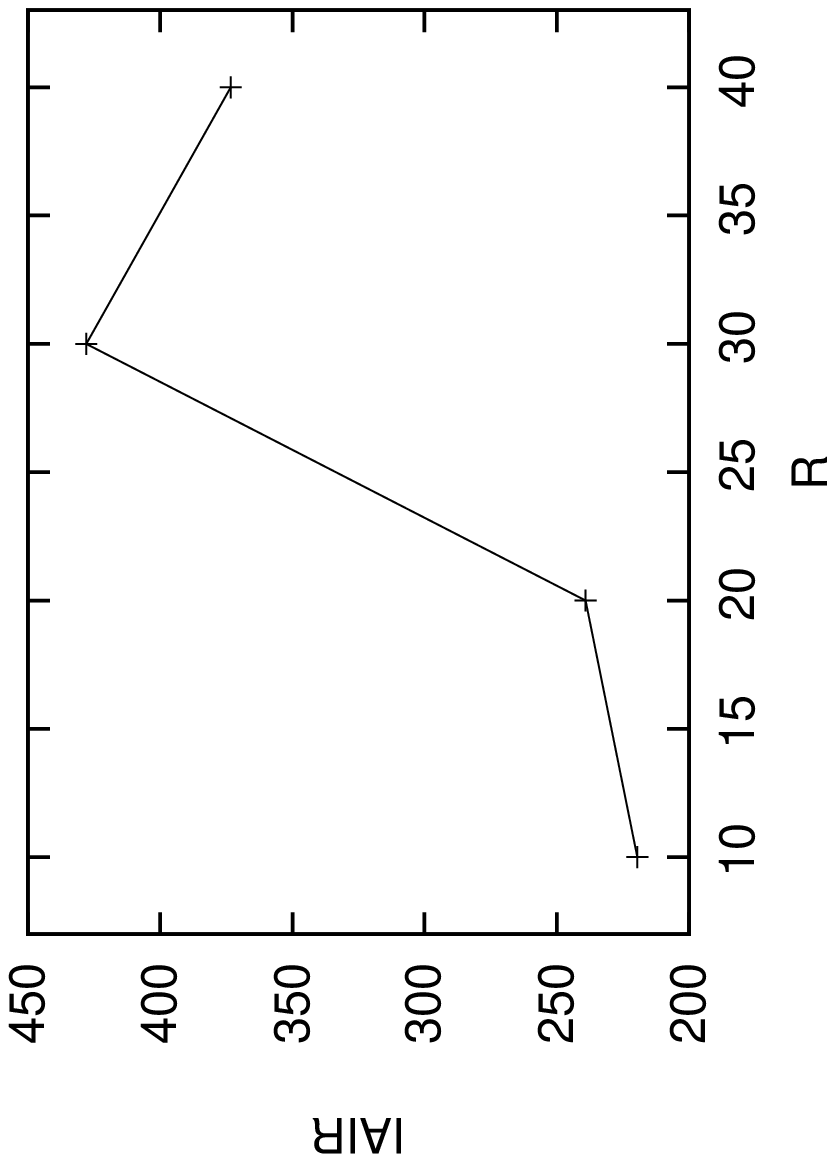}{(a)}\hspace*{-2mm}%
\FIGR{4}{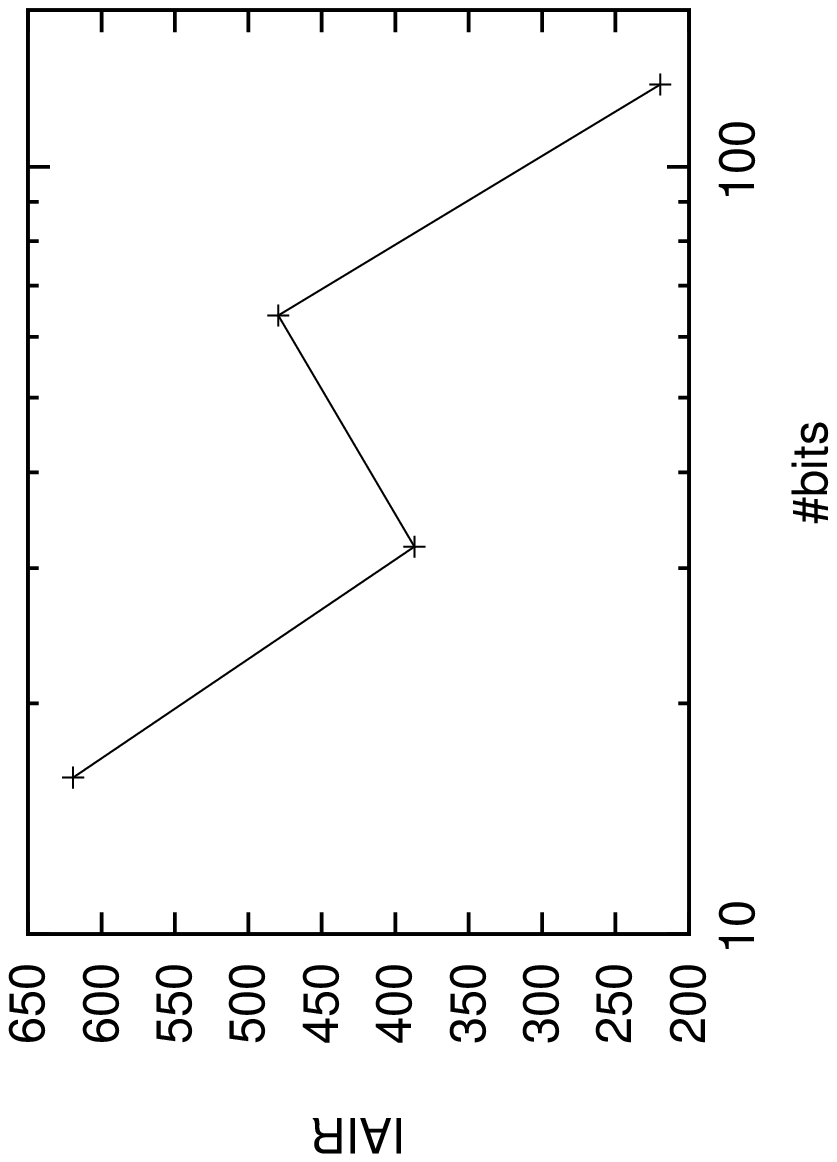}{(b)}\\
\FIGR{4}{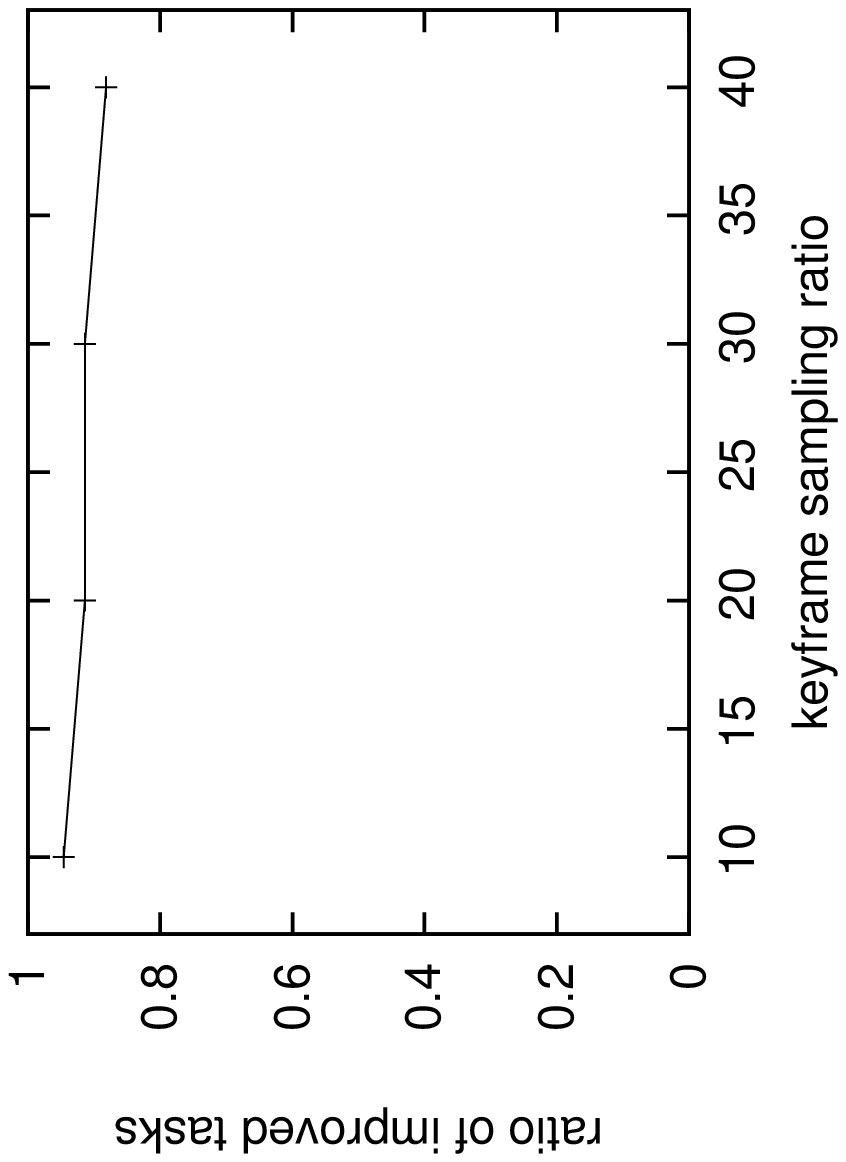}{(c)}\hspace*{-2mm}%
\FIGR{4.5}{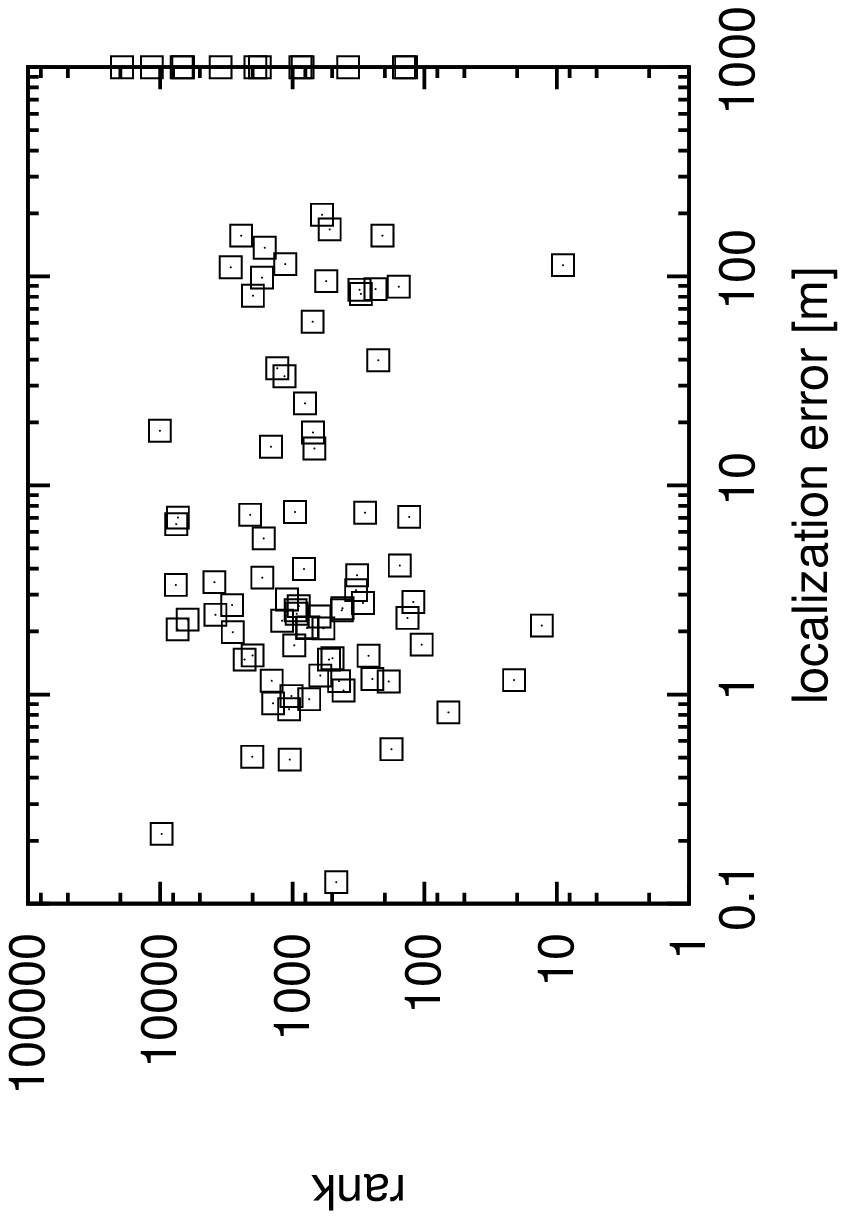}{(d)}\\
\FIGR{4}{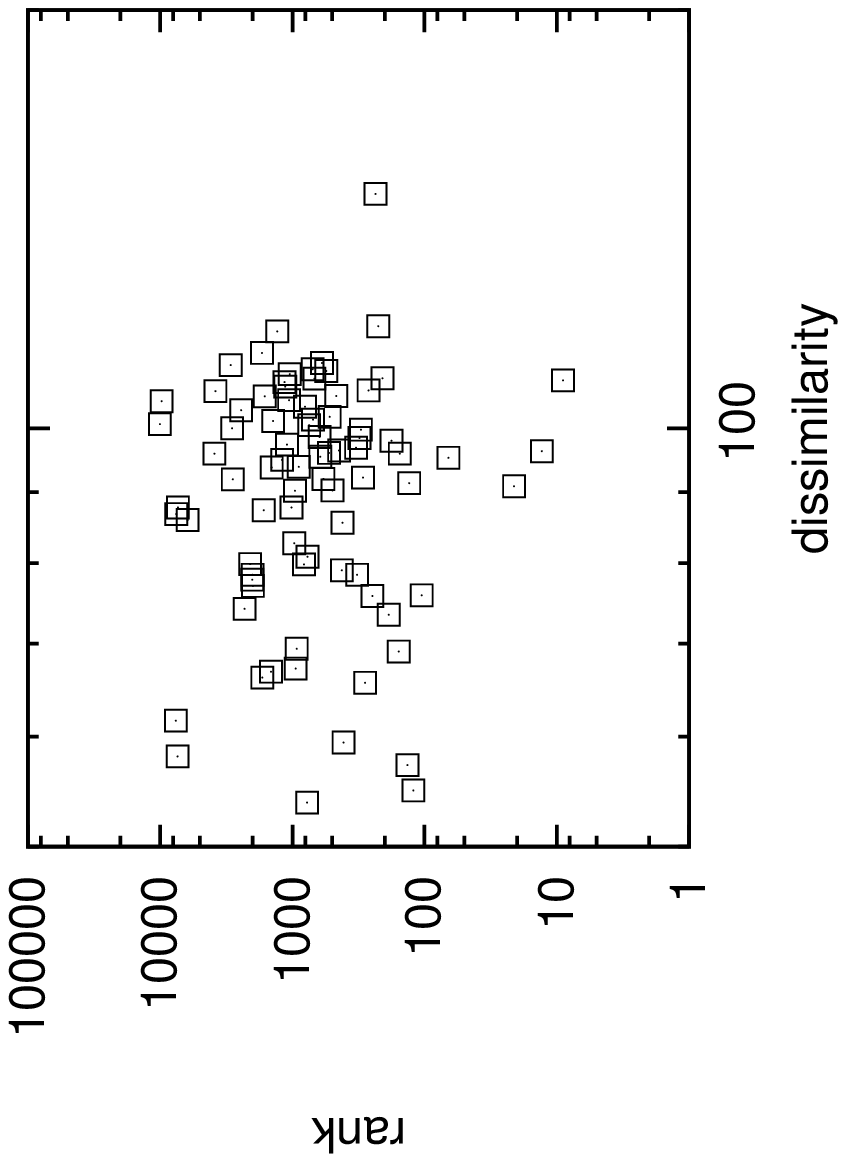}{(e)}
\end{minipage}
\caption{Change detection performance.
(a)
IAIR (Inverse average inverse rank) versus $R$.
(b)
IAIR versus \#bits.
  (c)
Effect of motion prior versus keyframe sampling ratio. The vertical axis shows ratio of tasks in which performace is better when motion prior is used than when not used. 
(d)
Influence of the ground-truth localization error. 
(e)
Influence of the dissimilarity estimated by the localization algorithm.      
}\label{fig:G}
\end{center}
\end{figure}
}

\newcommand{\figH}{
\begin{figure}[t]
  \begin{center}
 \FIG{8}{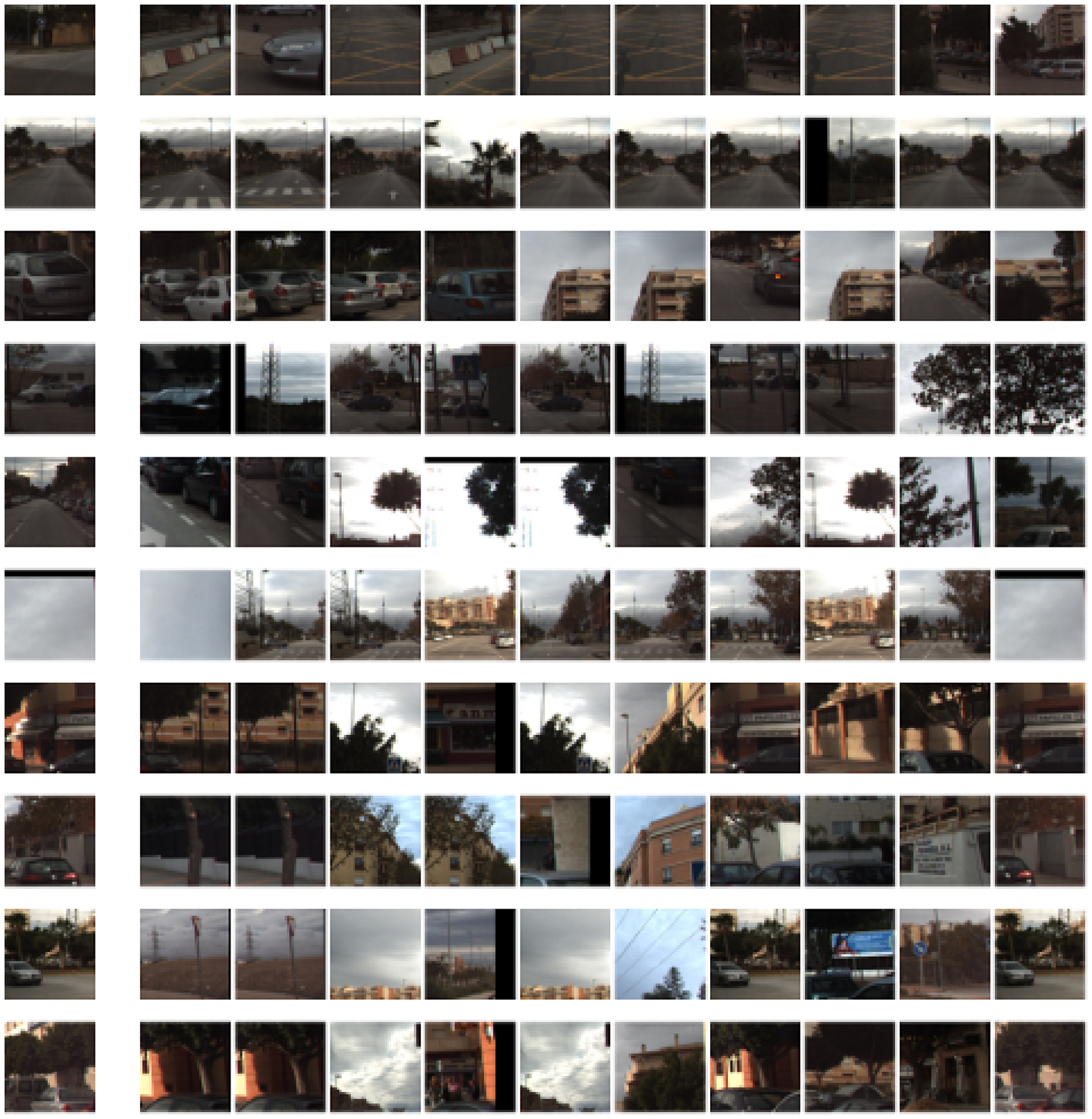}{}
\caption{Feature-level nearest neighbor search.
Each row corresponds to different features in different query images.
The left-most column shows a random instance of local feature in the query image, while the second to the last column shows the 1st, 2nd, ... nearest neighbor features.
For each panel, the local feature keypoint of interest is located at the center of the panel.
}\label{fig:H}
\end{center}
\end{figure}
}

\newcommand{\figI}{
\begin{figure}[t]
  \begin{center}
    \FIGR{8}{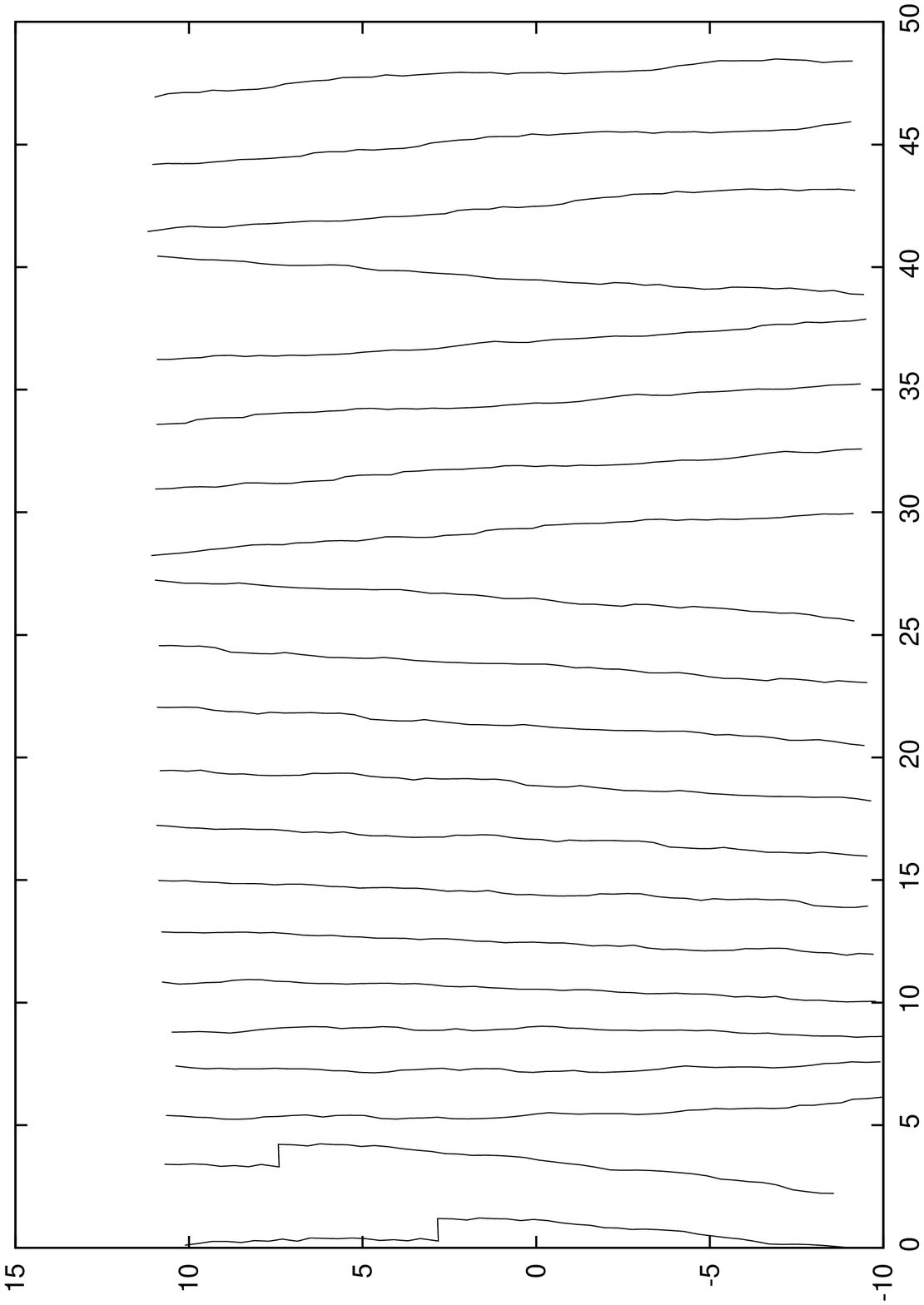}{}
    \FIGR{8}{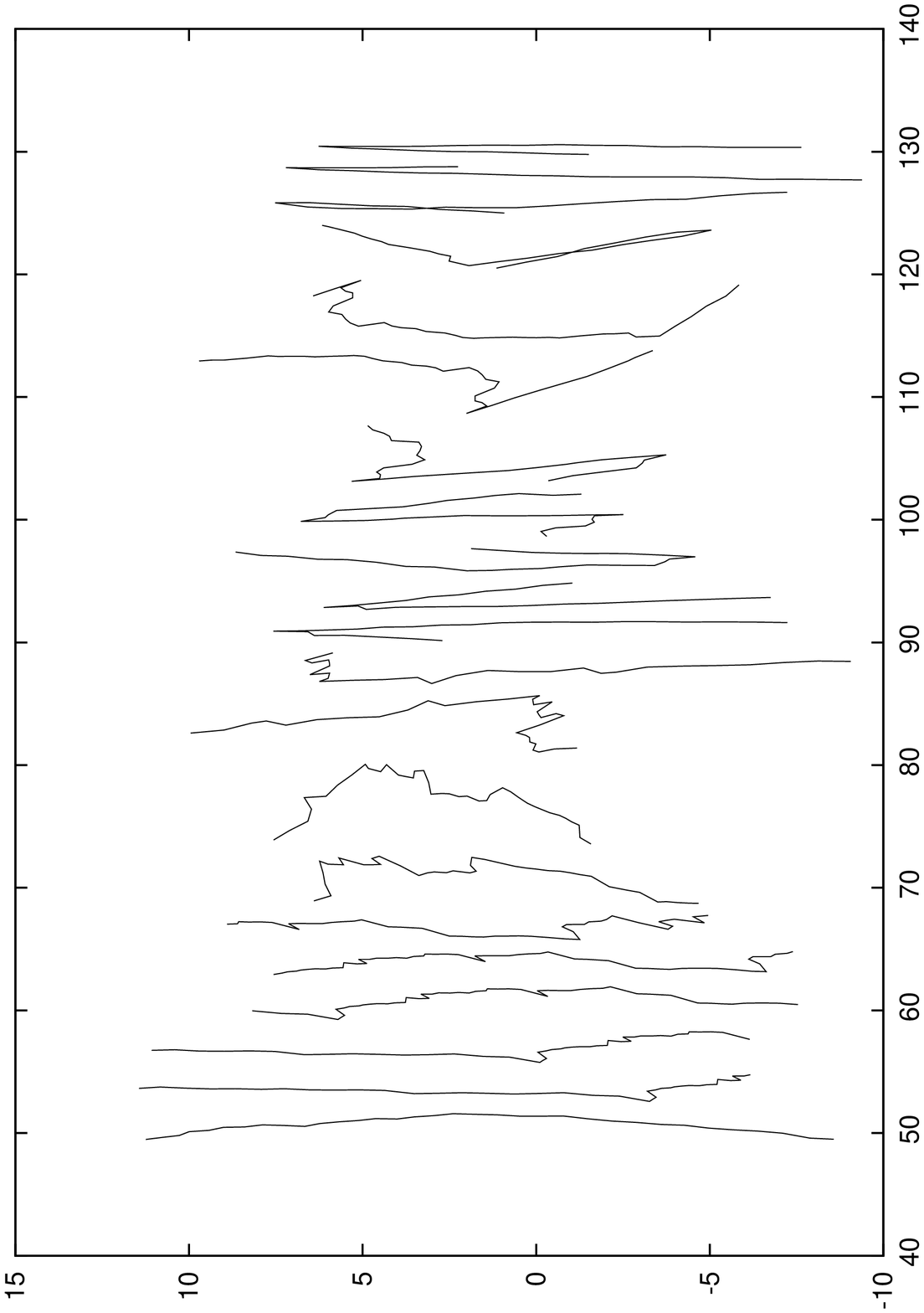}{}\vspace*{-3mm}\\
\caption{
  Training data for motion prior.
  Bird's eye views of several samples of the trajectories are shown in the graphs.
The top panel shows non-anomaly ego-motion samples in [m], while the bottom panel shows anomaly ego-motion samples.
}\label{fig:I}
\vspace*{-8mm}
  \end{center}
\end{figure}
}

\newcommand{\figJ}{
\begin{figure*}[t]
  \begin{flushleft}
~~~~\FIG{17}{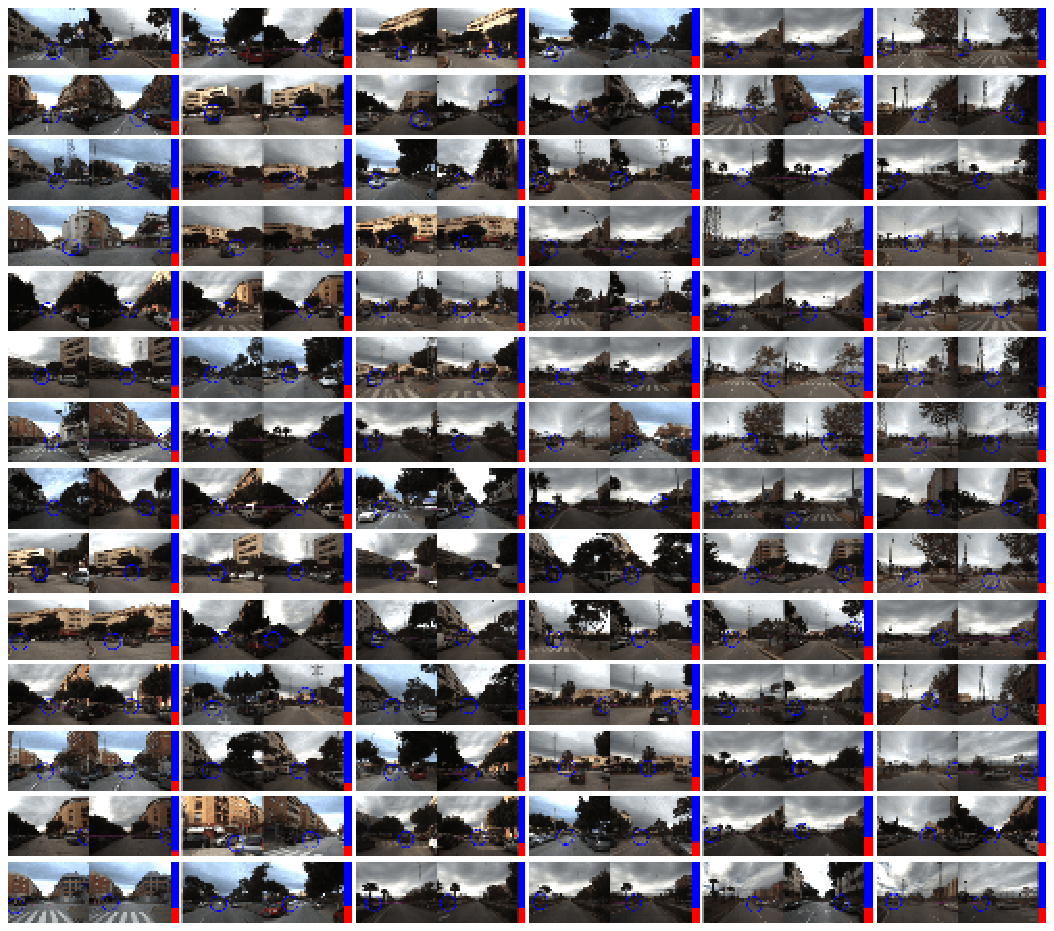}{}\vspace*{5mm}\\
\caption{Visualization of change detection.
  Each panel shows
  for each query image,
  the nearest neighbor feature pair 
  that is used to compute the likelihood of change.
  The blue circles and red colored bars show locations of the features and the likelihood of change.
  Each nearest neighbor feature pair is selected among
  every possible feature pair, one from the query image and the other from the top-$R$ ranked reference images.
}\label{fig:J}
\end{flushleft}
\end{figure*}
}

\renewcommand{\FIGpng}[5]{~ }

\author{Murase Tomoya ~~~~~~ Tanaka Kanji
\thanks{Our work has been supported in part by 
JSPS KAKENHI 
Grant-in-Aid for Young Scientists (B) 23700229,
and for Scientific Research (C) 26330297 
}
\thanks{The authors are with Graduate School of Engineering, University of Fukui, Japan.
{\tt\small tnkknj@u-fukui.ac.jp}}%
\vspace*{-5mm}}

\begin{document}

\title{\LARGE \bf
Change Detection under Global Viewpoint Uncertainty
}

\maketitle

\author{}

\begin{abstract}
  This paper addresses the problem of change detection from a novel perspective of long-term map learning. We are particularly interested in designing an approach that can scale to large maps and that can function under global uncertainty in the viewpoint (i.e., GPS-denied situations). Our approach, which utilizes a compact bag-of-words (BoW) scene model, makes several contributions to the problem:
  1) Two kinds of prior information are extracted from the view sequence map and used for change detection. Further, we propose a novel type of prior, called motion prior, to predict the relative motions of stationary objects and anomaly ego-motion detection. The proposed prior is also useful for distinguishing stationary from non-stationary objects. 
  2) A small set of good reference images (e.g., 10) are efficiently retrieved from the view sequence map by employing the recently developed Bag-of-Local-Convolutional-Features (BoLCF) scene model.
  3) Change detection is reformulated as a scene retrieval
  over these reference images to find changed objects
  using a novel spatial Bag-of-Words (SBoW) scene model.
  Evaluations conducted of individual techniques and also their combinations on a challenging dataset of highly dynamic scenes in the publicly available Malaga dataset verify their efficacy.
\end{abstract}

\section{Introduction}

Change detection is a key component for long-term map learning \cite{finman2013toward,arroyo2015towards,ott2012unsupervised} and has been attracting extensive research interest
in recent years \cite{wang2014cdnet}. In this paper, we address the problem of change detection from a novel perspective of long-term map learning. Given a single-view image acquired by a car-like robot, our approach localizes changed objects (e.g., other cars) with respect to a pre-built view sequence map (Fig. \ref{fig:A}). Specifically, we are interested in designing an approach that can scale to large maps and that can function under global uncertainty in the viewpoint (i.e., GPS-denied situations). Addressing
this problem at large	scale is of fundamental importance, particularly in the context of long-term map learning, owing to the
requirements of scalable map representation and global viewpoint localization.

\noeditage{
\figA
}

Thus far, the problem of change detection has been widely studied in the areas of computer vision and robot vision for various application domains including city model maintenance \cite{taneja2013city}, visual inspection \cite{stent2015detecting}, disaster monitoring \cite{gueguen2015large}, and patrol robots \cite{andreasson2007has}. The solutions include view registration \cite{fehr2016reshaping}, 3D line features \cite{eden2008using}, view synthesis \cite{taneja2013city}, occlusion reasoning \cite{silvadetecting}, and deep learning of patch-level similarity \cite{alcantarilla2016streetview}. 

Formulation as a scene comparison task, in which operations are carried out on a given pair of query and reference images, is common to the majority of these applications. To date, most of the state-of-the-art systems simply assume that relevant reference images are given, or rely on the availability of rough GPS information. However, providing relevant reference images is a non-trivial task in the case of long-term map learning. This is the main topic of our study.

This paper reformulates change detection as a scene retrieval task, in which both viewpoint localization and change detection are achieved by a scalable nearest neighbor algorithm with a compact bag-of-words (BoW) \cite{sivic2003video} scene model.

Our approach is related to previous work on scene retrieval but with key differences:
In contrast to visual place recognition or map relative viewpoint localization \cite{lowry2016visual}, we focus on retrieving not the whole image but a small object (i.e., the change) in the scene. Unlike particular object retrieval \cite{tolias2015particular}, we cannot assume the knowledge on where the target object (i.e., the change) is located in the input scene. Compared with common object discovery (COD) \cite{rubinstein2013unsupervised},
we need to identify not only common part but also changed part between scenes.

More specifically, our approach brings three contributions to the problem:

{\it 1) View Sequence Map as Prior:}
A view sequence map provides two kinds of prior information: appearance
prior and motion prior. The former can be naturally used as training data for vocabulary learning by the BoW model.
The latter provides prior for relative motions of stationary objects and anomaly ego-motion detection, which can be useful
for distinguishing stationary from non-stationary objects and also for evaluating the reliability of detection results.

{\it 2) Bag-of-Local-Convolutional-Features (BoLCF):}
Our viewpoint localization strategy is motivated by the recent success
in the local convolutional features from deep convolutional neural network (DCNN) \cite{zheng2016sift}
and their scalable BoW representation \cite{mohedano2016bags}.
We adopt the BoLCF technique to retrieve a small set of relevant reference images (such as 10) given a query image that is then used for change detection.

{\it 3) Spatial-Bag-of-Words (SBoW):}
Change detection is reformulated as a scene retrieval
over these reference images to find changed objects.
The results of viewpoint localization, motion prior, and appearance prior, are combined to compute the likelihood of change
using a novel spatial Bag-of-Words (SBoW) scene model.

We evaluated the effectiveness of individual techniques and also their combinations using a challenging dataset of highly
dynamic scenes in the publicly available Malaga dataset \cite{blanco2014malaga}. In addition to the above contributions, our experimental system can
also be viewed as a novel solution to the moving object detection (MOD) alternative task, which is complementary to existing
MOD approaches based on motion cues (e.g., motion segmentation, moving camera background subtraction) and appearance
cues (e.g., particular moving object recognition). 

In previous work, we investigated the problem of
global localization with change detection \cite{tanaka2004global},
cross-domain localization \cite{kanji2015cross},
and localization from images with small overlap \cite{kanji2016self}.
Our approach is
also inspired by existing techniques for self-localization in dynamic environments \cite{dellaert1999monte}, change detection \cite{pollard2007change}, motion anomaly detection \cite{roberts2009learning}, and tracking learning detection \cite{kalal2012tracking}.
However,
the problem of change detection under global viewpoint uncertainty has not yet been addressed in existing work.

\section{Problem}

\subsection{Dataset}

In contrast to previous change detection approaches, we focus on single-view recognition under highly dynamic scenes. This is more challenging than a typical scenario in which a complete 3D city model \cite{taneja2013city}  or a full 3D structure reconstruction
from multi-view images is used \cite{kovsecka2012detecting}. In our experiments, we utilized the publicly available Malaga dataset \cite{blanco2014malaga}, which
contains a set of view sequences for different robot trajectories. Although ground-truth GPS data and stereo images are also
available in this dataset, only a single-view image (left-eye view of the onboard stereo camera) is used by our change
detection algorithm. In the datasets, occlusion is severe in the scenes, and stationary objects can even have relative motions caused by the complex ego-motions of the robot-self, which makes our change recognition task a challenging one.

\subsection{Performance Index}

The performance of a change detection algorithm is evaluated over a set of query images. The output of a change detection
algorithm is a collection comprising the likelihood value for every local feature in every query image. We merge the outputs over all the query images and sort them in descending order of likelihood value. Then, the rank values of features that belong
to the ground-truth changed objects with respect to the sorted feature list is used as a measure for performance evaluation.
For the ground-truth changed objects, changed objects are manually annotated in the form of bounding boxes by comparing query and reference images. As the evaluation is based on ranking, a smaller value signifies better performance.
If multiple local features belong to the ground-truth bounding box, the rank value of the feature that is assigned the largest likelihood of change is used for the evaluation.

\subsection{Global Viewpoint Uncertainty}\label{sec:gvu}

In order to conduct change detection experiments under the challenging scenario of global viewpoint uncertainty, the a-priori
view sequence map is customized for individual query input images. Instead of using the full image dataset, a
subset of the images in the dataset whose time stamps are too close (closer than 400 frames) to the input image are considered nonmembers of the
view sequence. The customized view sequence map consists of a union of the image collections \#5, \#6, \#7, \#8 and \#10, minus the above mentioned subset of images. Note that this is a challenging setting, known as loop closure in the field of robotic mapping and localization, in which the viewpoint localization requires loop- closure detection \cite{lowry2016visual}. 

\section{Approach}

\subsection{Overview}

Our change detection task formulation follows a classical formulation, formulation as a regression problem, in which the
goal is to evaluate the likelihood of change for every local image feature in every query image. The standard solution for this
task is scene comparison between input and reference (i.e., mapped) scenes. As stated by many researchers, pre-registration
between the input and reference scenes is a necessary pre-processing step for reliable change detection \cite{radke2005image}. There are
primarily two solutions for the pre-registration. In one solution, availability of a complete 3D reference scene model
or full 3D reconstruction from a sequence of images (i.e., SLAM) or a collection of multi-view images (i.e., SfM) is assumed, and
the 3D model for the query is compared to those of the reference scenes \cite{kovsecka2012detecting}. The other solution is to directly compare the 2D input and reference images without assuming the availability of a 3D model. In this study, we employed the latter setting with global viewpoint uncertainty. This is a very challenging setting because the viewpoint uncertainty influences the pre-registration performance in a more direct manner than in the former case, in which the 3D model is available.

\noeditage{
\figB
}

More formally, we formulate the problem as follows. The basic idea is to predict the appearance $a$ and pose $p$ of an input local image feature, and then evaluate the difference in the observed feature $v=(a,p)$ from the prediction. The amount of difference can be viewed as anomalyness \cite{chandola2009anomaly}, or the likelihood of change \cite{radke2005image}. The prediction can be defined as the posterior distribution $P(a,p|I)$ of feature $v$ conditioned on the given input image $I$. The key observation is that the posterior distribution can be approximated by a set of features $V=\{v\}$ sampled from a subset of view images $I_1$, $\cdots$, $I_R$ in the prior view sequence map. It is natural to sample such a subset from the posterior distribution $P(v|I)$ of the current viewpoint $v$ given the input image $I$. A possible approach to compute this probability distribution is to utilize probabilistic localization algorithms such as Monte Carlo localization \cite{dellaert1999monte}. In this study, we simply approximated the sample set with a set of images in the view sequence map that are top-ranked by map relative localization, image retrieval, or place recognition subsystem. Let $I_i (i \in [1,R])$  denote the set of top-ranked $R$ reference images in the retrieval result. Let $W$ $=$$\{w_{ij}\}_{j=1}^{K_i}$ denote a feature set that consists of the feature $w_{ij} = (a_i, p_i)$ of each $i$-th top-ranked image. We approximate the likelihood of change as:
\begin{equation}
L(w) = \min_{i \in [1, R]} \min_{j \in [1,K_i]} D (w, w_{ij}).
\end{equation}
Note that we use $\min$ operation instead of the average operation. This is because we discovered that the average operation yields poor performance due to the fact that the majority of local features are contaminated by noise. In contrast, the $\min$ operation enables the filtering of such random noise because the chance of dissimilarity between input and random features being the min value is very low.

In the remainder of this section, we focus on the view sequence map of the target route and discuss the algorithm proposed for feature comparison (i.e., the $D$ function) and effective representation of features (i.e., $p$, $a$).
More specifically, we consider 
1) how to obtain the prior model for relative motions of 
stationary objects caused by ego-motion of the robot (\ref{sec:motion}), 
2) how to obtain the prior model for BoW appearance representation of objects used by both the viewpoint localization and change detection tasks (\ref{sec:appearance}), and 
3) how to utilize these prior models for change detection tasks. 

\noeditage{
\figI
}

\subsection{Motion Prior}\label{sec:motion}

Understanding the relative motions of stationary objects from visual experience along the target route is key to discriminating stationary and changed (dynamic/non-stationary) objects. In general, the relative motion of a static object between query and reference images can be explained by several factors, such as the robot's ego-motion, relative distance to the object, and the object's size and shape. In other words, if the relative motion of an object cannot be explained by these factors, the object can be changed (i.e., non-stationary or dynamic) object with high probability. We employ this idea to detect changed objects.

In the training phase, the view sequence map is the sole information source for learning. We learn the characteristics of the relative
motions of static objects from the available view sequence map (Fig. \ref{fig:B}). Conceptually, our approach is analogous to tracking learning detection (TLD) in the visual tracking community \cite{kalal2012tracking}. The algorithm consists of two steps. 1) In the first step, we extract KLT  features from each frame in the view sequence map and track them between adjacent frames. The trajectories of the KLT features during a unit length ego-motion can be viewed as motion features. For the ego-motion estimation, we use a monocular visual odometry with the five-point algorithm in \cite{nister2004efficient} in our own implementation. We simply approximate the trajectory as a 4D vector consisting of the trajectory's start and end points. 2) In the second step, we then construct a vocabulary of the 4D motion features. In the spirit of BoW vocabulary learning \cite{sivic2003video}, clustering over all the motion features over the entire view sequence map is performed and each
cluster is viewed as a motion word of our vocabulary.

Our clustering algorithm finds the largest clusters and assigns motion word to each of them. The basic idea is to use the consistency check to filter out outlier motion features contaminated by noise and to discover inlier motion features to be used as the motion word candidates. It randomly samples $N=$10,000 motion features to construct a database of motion features and then performs an $N$ set of one-nearest neighbor retrieval over the remaining $N-1$ features, using each feature as a query. For consistency check, the retrieved feature is checked if the query feature is also the one-nearest neighbor feature for the retrieved feature. If this condition is true, the query feature is considered as passing the consistency check. The above procedure comprising sampling, NN retrieval, and consistency checking is iterated for 100 different sets of random databases. Then, 1,000 features that have passed the largest number of consistency checks are output as the 1,000 motion words.

We also introduce an anomaly ego-motion detection strategy from monocular visual odometry. This strategy is motivated by observation that relative motion measurement is not always reliable. Specifically, it tends to be reliable while the robot is in non-anomaly ego-motion such as straight-line motion, but becomes unstable when the robot is in anomaly ego-motion such as curved motion or slip motion.

For each frame, we monitor the ego-motion measurement and if the curvature of the ego-motion trajectory exceeds a predefined threshold $T_c=5$ deg, we simply do not use the relative motion measurement for that frame. For a given trajectory length $L$, the curvature is defined as the deviation of exemplar
directions [rad], computed from a pairing of the start and end points. For these start and end points, we use the $i$-th viewpoint and $i+L/2$-th viewpoint, respectively, for each different $i$ in $[0,L/2-1]$.
Fig. \ref{fig:I} visualizes samples of tracking data for non-anomaly and anomaly ego-motion classes, as discussed in \ref{sec:motion}.

In order to reduce the storage cost for the view sequence map, we also consider the keyframe selection task. Keyframe selection involves finding representative frames in the view sequence map. Once the keyframe set is determined, the change detection task approximates each input image by its nearest neighbor keyframe in terms of the view ID. In our baseline strategy, we sample a keyframe every 10 frames in the view sequence map.

\subsection{Appearance Prior}\label{sec:appearance}

We employ the BoW representation for appearance features that are used for change detection. Two types of BoW representations with different levels of trade-offs between compactness and discriminativity and between accuracy and robustness are used for the two different
tasks: map relative localization and change detection.

\subsubsection{Bag of Local Convolutional Features}

BoW representation compactness and discriminativity are basic requirements for viewpoint localization \cite{lowry2016visual}. In general, compactness is realized by limiting the number of visual words per mapped image. Conversely, discriminativity depends strongly on the choice of the local feature descriptor. Considering these requirements, we employ local convolutional features with BoW representation---a technique developed in the image retrieval community \cite{mohedano2016bags}. The basic idea of this technique is to pre-train a deep convolutional neural network on big data (e.g., imagenet), and then view responses from its convolutional layer as a grid of high-dimensional (e.g., 256-dim) local feature descriptors. The technique has been found to be computationally efficient and competitive with other state-of-the-art scene matching and retrieval algorithms \cite{mohedano2016bags}. In our approach, we adopt Caffenet and use its last convolutional layer as a size 169 set of 256-dim local feature descriptors. A fine vocabulary with size 1M is learned from an independent dataset and used to convert every local convolutional feature to a 20-bit code. As several researchers have stated, and as also discovered by us in our preliminary experiments, a key limitation of such a fine vocabulary is that sensitivity increases in the vector quantization. To address this issue, we employ asymmetric feature comparison using the NBNN distance metric, as detailed in our previous study \cite{kanji2015cross}. In this method, the distance between a query image's feature set $I^{query}=\{f\}$ and a reference image's word set $I^{reference}=\{w\}$ is computed as follows:
\begin{equation}
D_{NBNN}(I^{query}, I^{reference}) 
= 
\sum_f
\min_{w} | f - \bar{f}(w) |,
\end{equation}
where
$\bar{f}$
is a function
that returns the exemplar feature
corresponding to an input visual word $w$.
As also shown in our previous work \cite{kanji2016self},
viewpoint localization using the NBNN distance metric
is stable and works even 
when there is no common visual words between query and reference images.

\figH

\subsubsection{Bag of Binarized SIFT Words}

Feature representation accuracy and robustness are basic requirements for change detection \cite{radke2005image}. In general, accuracy is realized by employing a fine vocabulary. On the other hand, robustness depends on the choice of local feature descriptor.
We employ the combination of
harrislaplace detector 
and 
SIFT feature descriptor, which has proven to be robust in various change detection tasks \cite{kovsecka2012detecting}.
A random projection technique as in \cite{kanji2016self} is employed as a dictionary to translate each SIFT vector to a more compact
$B=128$-bit binary code, to obtain a BoW representation in the split of bag-of-binary-words \cite{arroyo2015towards}. Our fine vocabulary requires a number of bits per local feature descriptor and the database of BoW representations cannot operate in main memory. Fortunately, we can expect that the map relative localization provides a sufficiently small set of $R=10$ reference image candidates, which requires a reasonably small space per query image. Fig. \ref{fig:H} shows random examples of nearest neighbor search. The examples include dynamic objects such as cars, and static objects such as road, wall, shop, and sky. It can be seen that similar objects are successfully found in the examples shown.

\subsubsection{Nearest Neighbor Anomaly Detection}

The results of above tasks---map relative localization, motion prior, and appearance prior---are combined to compute the likelihood of change. Incorporating motion prior to evaluate the likelihood of change is a non-trivial task. In our SBoW approach, we represent a hypothesized motion of a query local feature of interest by a 4D vector $(x^q, y^q, x^r, y^r)$, where $x^q, y^q$ represents the 2D pixel location of a query local feature, and $x^r, y^r$ represents the 2D pixel location of its nearest neighbor reference local feature.
Then, we test if the distance between the query and its nearest neighbor motion feature
in the 4D motion feature space is greater than a pre-defined threshold $T_m=10$. 
If it is greater, the query motion feature is considered as belonging to the anomaly motion class. Our criterion for evaluating the likelihood of change of a given query local feature $f$ is in the form:
\begin{equation}
L(f) = 
\min_{\hat{f}\in A(f)} \left(~ | f - \hat{f} |~ +~ M(f) | f - \hat{f} | ~ \right),
\end{equation}
where $A(f)$ is a function that returns nearest neighbor $K=10$ features in the reference image, and $M(f)$ is a function that takes one if the motion feature of $f$ belongs to the anomaly motion class, and zero otherwise.

\section{Experiments}

Experiments were conducted to validate our approach on several change detection tasks. Their results indicate that the proposed algorithm is memory efficient, performs well, and scales to large maps. We also compared our approach with a baseline method that does not use the motion prior, and also analyzed the sensitivity of the approach to viewpoint uncertainty.

Fig. \ref{fig:C} gives 
a bird's eye view of the viewpoint trajectory of the mapper robot of the view sequence maps used in the experiments.
We used 
sequences \#5, \#6, \#7, \#8, and \#10 in the Malaga dataset.
This is because they are reasonably long
sequences and, more importantly, they contain the loop-closure situations, which correspond to the map relative localization under global viewpoint uncertainty.
We also used sequence \#9 (length 1,018) as 
the training data
for learning motion prior and appearance prior,
in the procedure described in subsections \ref{sec:motion} and \ref{sec:appearance}.
Each image in the dataset is sized $1,024\times 768$.
Sequences \#5, \#6, \#7, \#8, and \#10 
contain 4,816, 4,618, 2,122, 10,026, and 17,310 images, respectively.

We created the test set considering two requirements. 1) The timestamps of all the images in the view sequence map should not be close (closer than 400 frames) to that of the query image. This setting, comparison between new and old images, is common to many change detection applications. This requirement is met by the procedure described in \ref{sec:gvu}. 2) The query image's view should partially overlap at least one reference image in the view sequence map. Otherwise, change detection algorithms can fail badly as pre-registration usually requires partial view overlap between query and reference images. To meet this requirement, we select query images in the test set such that each query image's viewpoint is located on the mapper robot's viewpoint trajectory of the view sequence map. Note that this is analogous to a situation called loop-closure situation in the field of robotic mapping and localization. We approximately uniformly sampled from the images that met the above two requirements and obtained a set of 94 pairings of query image and view sequence map.

\noeditage{
\figC
}

\noeditage{
\figG
}

\noeditage{
\figF
}

\noeditage{
\figD
}

\noeditage{
\figJ
}

Fig. \ref{fig:G} shows change detection performance.
It can be seen that
for most of the cases considered here,
better performance is obtained
when motion prior is used
than
when not used.
Change detection performance tended to be higher
when
the number of bits per binary word
is relatively high.
It can be also seen
that change detection was
successful for
wide range of localization error,
which
is defined as the Euclidean distance between the ground-truth viewpoint of the query image and that of the reference image
that is top-ranked by the map relative localization.
Our change detection algorithm is frequently successful even when the localization error is large (e.g., 100 m).
In fact, success in global viewpoint localization is not a necessary condition for change detection.
The map relative localization often found reference images with a similar landscape to that in the query
image, and then our change detection algorithm was able to identify change using the difference in the appearance and location of the changed object as a cue. In summary, we obtained the following results.
\begin{enumerate}
\item
$B=128$ binary code was necessary for reliable change detection.
\item
Top $R=10$ -ranked reference image set was already sufficient for viewpoint localization.
\item
In 94 \% of the tasks, motion prior was effective to improve change detection performance.
\end{enumerate}

Fig. \ref{fig:F} shows map relative localization results for individual query images.
It can be seen that
the top-$R$ ranked reference image set is reliable when $R$ is equal or larger than 5 in these cases considered here.

Fig. \ref{fig:D} shows change detection results for individual query images.
It can be seen that the method that combines the appearance and motion priors
more frequently assigns a small rank value to 
the ground-truth changed features.

Fig. \ref{fig:J} shows examples of change detection.
It can be seen that change detection was frequently successful
even when
the query image was not well registered against the reference image,
which corresponds to
the conditions of local or global viewpoint uncertainty.
It can be also seen that
due to the dynamic nature of the traffic environments,
the algorithm was often confused by
visually similar but different dynamic objects the robot encountered.
It is natural
that change detection
becomes a difficult task
when the scene contains confusing changed objects.
Note that
such changed objects
can be detected and removed from the view sequence map
to some extent
in the TLD framework.
Despite the difficulty,
our approach was more frequently successful by
making use of the appearance and motion prior as a cue.

\section{Conclusions}

In this paper, we
addressed a novel problem of
change detection under global viewpoint uncertainty. For compactness and efficiency, the proposed method employs two types of BoW scene models---BoLCF and SBoW---using the view sequence map as a prior. In addition, we proposed a novel prior, called motion prior, to represent the difference in the location of the local feature keypoint between query and reference images. Our BoW retrieval formulation enables scalable change detection even when there is minimal view overlap between query and reference images. As the approach is quite simple, it is applicable to diverse change detection applications.

\bibliographystyle{IEEEtran}
\bibliography{mod}

\end{document}